\documentclass[runningheads]{llncs}
\usepackage{graphicx}
\usepackage{arabtex}
\usepackage{utf8}
\usepackage[utf8]{inputenc}
\usepackage[arabic,USenglish]{babel}
\begin{document}
\title{Learning Word Representations for Tunisian Sentiment Analysis \thanks{Supported by iCompass}}
%
%
\author{Abir	Messaoudi \and
Hatem	Haddad  \and
Moez Ben HajHmida \and
Chayma	Fourati \and
Abderrazak	Ben Hamida
}
\authorrunning{Abir	Messaoudi et al.} 

%
\institute{iCompass, Tunisia \\
\email{\{abir,hatem,moez,chayma,razzak\}@icompass.digital}\\
\url{https://www.icompass.tn}}
\maketitle              
\begin{abstract}
Tunisians on social media tend to express themselves in their local dialect using Latin script (TUNIZI). This raises an additional challenge to the process of exploring and recognizing online opinions. To date, very little work has addressed TUNIZI sentiment analysis due to scarce resources for training an automated system. In this paper, we focus on the Tunisian dialect sentiment analysis used on social media. Most of the previous work used machine learning techniques combined with handcrafted features. More recently, Deep Neural Networks were widely used for this task, especially for the English language. In this paper, we explore the importance of various unsupervised word representations (word2vec, BERT) and we investigate the use of Convolutional Neural Networks and Bidirectional Long Short-Term Memory. Without using any kind of handcrafted features, our experimental results on two publicly available datasets~\cite{tunizi}~\cite{tsac} showed  comparable performances to other languages.

\end{abstract}
\section{Introduction}

Since the end of the twentieth century and the spread of mobile communication technologies in the Arab world, youth, in particular, have developed a new chat alphabet to communicate more efficiently in informal Arabic. Because most media and applications initially did not enable chatting in Arabic, these Arab speakers resorted to what is now commonly known as "Arabizi". In~\cite{TAL}, Arabizi was defined as the newly-emerged Arabic variant written using the Arabic numeral system and Roman script characters. With the widespread use of social media worldwide in more recent years, Arabizi emerged as an established Arabic writing system for mobile communication and social media in the Arab world.

Compared to the increasing studies of sentiment analysis in Indo-European languages, similar research for Arabic dialects is still very limited.\ This is mainly attributed to the lack of the needed good quality Modern Standard Arabic (MSA) publicly-available sentiment analysis resources in general~\cite{9035299}, and more specifically dialectical Arabic publicly-available resources.\ Building such resources involves several difficulties in terms of data collection and annotation, especially for underrepresented Arabic dialects such as the Tunisian dialect. Nevertheless, existing Tunisian annotated datasets~\cite{medhaffar-etal-2017-sentiment} focused on code-switching datasets written using the
Arabic or the Romanized Alphabet. The studies on these datasets applied off-the-shelf models
that have been built for MSA on a dataset of Tunisian Arabic. An intuitive solution is to translate Tunisian Romanized Alphabet into Arabic Script~\cite{10.1145/3364319}. This approach suffers from the need for a parallel Tunisian-Arabic text corpus, the low average precision performances achieved and the irregularity of the words written.

Using a model trained on Modern Standard
Arabic sentiment analysis data
and then applying the same model on dialectal sentiment analysis data, does not
produce good performances as shown in~\cite{qwaider-etal-2019-modern}. This suggests that MSA
models cannot be effective when applied to dialectical Arabic. There is, thus, a growing need for the
creation of computational resources, not only for
MSA but also for dialectical Arabic. The same situation
holds when one tries to use computational
resources used for a specific dialect of Arabic with
another one.

To the best of our knowledge, this is the first study on sentiment analysis TUNIZI Romanized Alphabet. \ This could be deduced in the next sections where we will present TUNIZI and the state-of-the-art of Tunisian sentiment analysis followed by our proposed approach, results and discussion before conclusion and future work.

\section{TUNIZI}
Tunisian dialect, also known as ``Tounsi'' or ``Derja'', is different from Modern Standard Arabic. In fact, Tunisian dialect features Arabic vocabulary spiced with words and phrases from Tamazight, French, Turkish, Italian and other languages~\cite{tounsi}. 

Tunisia is recognized as a high contact culture where online social networks play a key role in facilitating social communication~\cite{emerging}. In~\cite{fourati}, TUNIZI was referred to as "the Romanized alphabet used to transcribe informal Arabic for communication by the Tunisian Social Media community". To illustrate more, some examples of TUNIZI words translated to MSA and English  are presented in Table~\ref{table1}.

\begin{table}
\centering
\caption{Examples of TUNIZI common words translated to MSA and English.}\label{table1}
\begin{tabular}{|l|l|l|l|}
\hline
TUNIZI  & MSA translation & English translation \\
\hline
3asslema       & \setcode{utf8}\<مرحبا> & Hello \\
chna7welek      &   \setcode{utf8}\<كيف حالك >    & How are you \\
sou2el           &  \setcode{utf8}\<سؤال> & Question \\
5dhit           &  \setcode{utf8}\<أخذت> & I took \\
\hline
\end{tabular}
\end{table}

Since some Arabic characters do not exist in the Latin alphabet, numerals, and multigraphs instead of diacritics for letters, are used by Tunisians when they write on social media. For instance, "ch" is used to represent the character  \textRL{\foreignlanguage{arabic}{ش}}. An example is the word \textRL{\foreignlanguage{arabic}{شرير}}\footnote{Wicked} represented as "cherrir" in TUNIZI characters.

After a few observations from the collected datasets, we noticed that Arabizi used by Tunisians is slightly different from other informal Arabic dialects such as Egyptian Arabizi. This may be due to the linguistic situation specific to each country. In fact, Tunisians generally use the french background when writing in Arabizi, whereas, Egyptians would use English. For example, the word \textRL{\foreignlanguage{arabic}{مشيت}} would be written as "misheet" in Egyptian Arabizi, the second language being English. However, because the Tunisian's second language is French, the same word would be written as "mchit". 

In Table~\ref{arabizi},  numerals and multigraphs are used to transcribe TUNIZI characters that compensate the the absence of equivalent Latin characters for exclusively Arabic arabic sounds. They are represented with their corresponding Arabic characters and Arabizi characters in other countries. For instance, the number 5 is used to represent the character  \textRL{\foreignlanguage{arabic}{خ}} in the same way as the multigraph "kh". For example, the word "5dhit" is the representation of the word \textRL{\foreignlanguage{arabic}{أخذت}} as shown in Table~\ref{table1}. Numerals and multigraphs used to represent TUNIZI are different from those used to represent Arabizi. As an example, the word \textRL{\foreignlanguage{arabic}{غالية}}\footnote{Expensive.} written as "ghalia" or "8alia" in TUNIZI corresponds to "4'alia" in Arabizi.

\begin{table}
\centering
\caption{Special Tunizi characters and their corresponding Arabic and Arabizi characters.}\label{arabizi}
\begin{tabular}{|l|l|l|l|}
\hline
 Arabic   &  Arabizi   &  TUNIZI   \\
\hline

\textRL{\foreignlanguage{arabic}{ح}} & 7 & 7 \\
\textRL{\foreignlanguage{arabic}{خ}} & 5 or 7' & 5 or kh\\
\textRL{\foreignlanguage{arabic}{ذ}} & d' or dh & dh\\
\textRL{\foreignlanguage{arabic}{ش}} & \$ or sh & ch\\
\textRL{\foreignlanguage{arabic}{ث}} &  t' or th or 4 & th\\
\textRL{\foreignlanguage{arabic}{غ}} & 4' & gh or 8\\
\textRL{\foreignlanguage{arabic}{ع}} & 3 & 3\\
\textRL{\foreignlanguage{arabic}{ق}} & 8 & 9\\

\hline
\end{tabular}
\end{table}

~\cite{younes} mentioned that 81\% of the Tunisian comments on Facebook used Romanized alphabet. In~\cite{abidi}, a study was conducted on 1,2M social media Tunisian comments (16M words and 1M unique words) showed that 53\% of the comments used Romanized alphabet while 34\% used Arabic alphabet and 13\% used script-switching. The study mentioned also that 87\% of the comments based on Romanized alphabet are TUNIZI, while the rest are French and English. 
In~\cite{survey}, a survey was conducted  to address the availability of Tunisian Dialect datasets. The authors concluded that all existing Tunisian datasets are using the Arabic alphabet or a code-switching script and that there is a lack of Tunisian Romanized alphabet annotated datasets, especially datasets to address the sentiment analysis task.  

~\cite{parallelcorpus} presented a multidialectal parallel corpus of five Arabic dialects: Egyptian, Tunisian, Jordanian, Palestinian and Syrian to identify similarities and possible differences among them. The overlap coefficient results, representing the percentage of lexical overlap between the dialects, revealed that the Tunisian dialect has the least overlap with all other Arabic dialects. On the other hand, because of the specific characteristics of the Tunisian dialect, it shows
phonological, morphological, lexical, and syntactic
differences from other Arabic dialects such that Tunisian words might infer
different syntactic information across different
dialects; and consequently different meaning and sentiment polarity than Arabic words.

These results highlight the problem that the Tunisian Dialect is a low resource language and there is a need to create Tunisian Romanized alphabet datasets for analytical studies. Therefore, the existence of such a dataset would fill the gap for research purposes, especially in the sentiment analysis domain.

\section{Tunisian Sentiment Analysis}

In~\cite{Karmani}, a lexicon-based sentiment analysis system was used to classify the sentiment of Tunisian tweets. The author developed a Tunisian morphological analyzer to produce linguistic features and achieved an accuracy of 72.1\% using the small-sized TAC dataset (800 Arabic script tweets).

~\cite{sayadi} presented a supervised sentiment analysis system for Tunisian Arabic script tweets. With different bag-of-word schemes used as features, binary and multiclass classifications were conducted on a Tunisian Election dataset (TEC) of
3,043 positive/negative tweets combining
MSA and Tunisian dialect. The support vector machine was found of the best results for binary classification with an accuracy of 71.09\% and an F-measure of 63\%.

In~\cite{medhaffar-etal-2017-sentiment}, the doc2vec algorithm was used to produce document embeddings of Tunisian Arabic and Tunisian Romanized alphabet comments. The generated embeddings were fed to train a Multi-Layer Perceptron (MLP) classifier where both the achieved accuracy and F-measure values were 78\% on the TSAC (Tunisian Sentiment Analysis Corpus) dataset. This dataset combines 7,366 positive/negative Tunisian Arabic and Tunisian Romanized alphabet Facebook comments. The same dataset was used to evaluate Tunisian code-switching sentiment analysis in~\cite{Jerbi} using the LSTM-based RNNs model reaching an accuracy of 90\%.

In~\cite{haddad2018}, authors conducted a study on the impact on
the Tunisian sentiment classification performance when it is
combined with other Arabic based preprocessing tasks (Named Entities tagging, stopwords removal, common emoji recognition, etc.). A lexicon-based approach and the support vector machine model were used to evaluate the performances on the above-mentioned datasets (TEC and TSAC datasets).

In order to avoid the hand-crafted features labor-intensive task, syntax-ignorant n-gram embeddings representation composed and learned using an unordered composition function and a shallow neural model was proposed in~\cite{syntax}. The proposed model, called Tw-StAR, was evaluated to predict the sentiment on five Arabic dialect datasets including the TSAC dataset ~\cite{medhaffar-etal-2017-sentiment}.

We observe that none of the existing Tunisian sentiment analysis studies focused on the Tunisian Romanized alphabet which is the aim of this paper.

Recently, a sentiment analysis Tunisian Romanized alphabet  dataset was introduced in ~\cite{tunizi} as "TUNIZI". The dataset includes more than 9k Tunisian social media comments written only using Latin script was introduced in~\cite{fourati}. The dataset was annotated by Tunisian native speakers who described the comments as positive or negative.  The dataset is balanced, containing 47\% of positive comments and 53\% negative comments.

\begin{table}
\centering
\caption{Positive and negative TUNIZI comments translated to  English.}\label{table2}
\begin{tabular}{|l|l|l|}
\hline
TUNIZI &  English & Polarity\\
\hline
enti ghalia benesba liya & You are precious to me & Positive \\
nakrhek 5atrek cherrir & I hate you because you are wicked & Negative\\
\hline
\end{tabular}
\end{table}

Table~\ref{table2} presents examples of TUNIZI dataset comments with the translation to English where the first comment was annotated as positive and the second as negative.

TUNIZI dataset statistics, including the total number of comments, number of positive comments, number of negative comments, number of words, and number of unique words are stated in Table~\ref{table3}.

\begin{table}
\caption{TUNIZI and TSAC-TUNIZI datasets statistics.}\label{table3}
\begin{tabular}{|l|l|l|l|l|l|l|l|}
\hline
Dataset & \#Words & \#Uniq Words & \#Comments & \#Negative & \#Positive & \#Train & \#Test \\
\hline
TUNIZI& 82384  & 30635 & 9911 & 4679 & 5232 & 8616 & 1295 \\
TSAC-TUNIZI &  43189  & 17376 &9196 & 3856 &  5340 &7379 & 1817 \\
\hline
\end{tabular}
\end{table}

For the purpose of this study, we filtred the TSAC dataset~\cite{medhaffar-etal-2017-sentiment} to keep only Tunisian Romanized comments. The new dataset that we called TSAC-TUNIZI includes 9196 comments. The dataset is not balanced as the majority of the comments are positive (5340 comments). TSAC-TUNIZI dataset statistics are presented in Table~\ref{table3}. 
\section{Proposed Approach}
 
In this section, we describe the different initial representations used, the hyper-parameters' values and the classifiers' architectures. 

\subsection{Initial Representations}

In order to evaluate Tunisian Romanized sentiment analysis, three initial representation were used: word2vec, frWaC and multilingual BERT (M-BERT).
\begin{itemize}
    \item Word2vec~\cite{word2vec}: Word-level embeddings were used in order to represent words in a high dimensional space. We trained a word2vec 300-vector on the TUNIZI dataset.

\item frWaC~\cite{fauconnier}: Word2vec pretrained on 1.6 billion French word dataset constructed from the Web. The use of the French pretrained word embedding is justified by the fact that most of the Tunizi comments present either French words or words that are inspired by French but in the Tunisian dialect. Examples of TUNIZI comments containing code switching are shown in Table~\ref{table4}.

\begin{table}
\caption{Examples of Tunizi comments switching-code.}\label{table4}
\begin{tabular}{|l|l|l|}
\hline
Comment & \ Languages \\
\hline
\textbf{je suis d'accord avec vous} fi hkeyet ennou si 3andha da5l fel hkeya bouh. & Tun, \textbf{Fr}\\
\textbf{good} houma déjà 9alou fih 11 7al9a. & Tun, \textbf{Eng}\\
t5afou mnha \textbf{vamos} el curva sud y3ayech weldi bara a9ra.& Tun, \textbf{Spa} \\
\hline
\end{tabular}
\end{table}

\begin{figure}
\includegraphics[width=\textwidth]{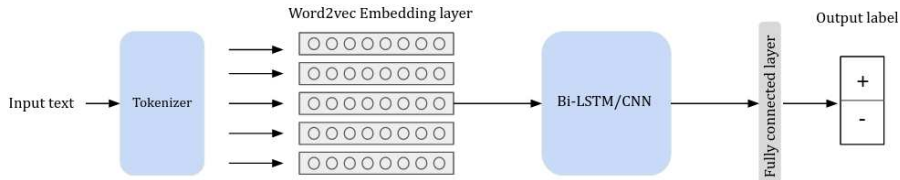}
\caption{Global architecture of the first proposed model} \label{fig1}
\end{figure}

\item Multilingual BERT (M-BERT)~\cite{devlin-etal-2019-bert}:
We decided to use the Bidirectional Encoder Representations from Transformers (BERT)  as a contextual language model in its multilingual version as an embedding technique, that  contains  more  than  10  languages including English, French and Arabic. Using the BERT tokenizer, we map words to their indexes and representations in the BERT embedding matrix. These representations are used to feed the classification models.
\begin{figure}
\includegraphics[width=\textwidth]{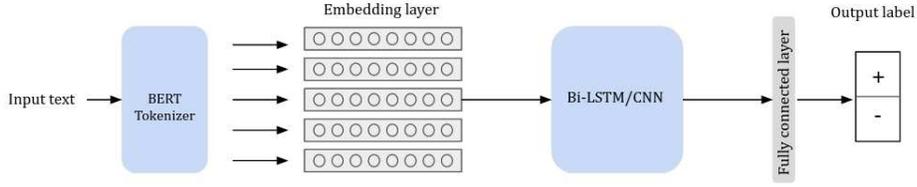}
\caption{Global architecture of the second proposed model} \label{fig2}
\end{figure}
\end{itemize}

\subsection{Classifiers}
Based on the state of the art, RNNs and CNNs are considered to be more efficient for the sentiment analysis (SA) task. In this work, we experimented two classifiers, a Convolutional Neural Network (CNN) with different sizes of filters and a Bidirectional Long Short-Term Memory (Bi-LSTM), a variant of RNNs. As a consequence, we ended up with two different models. The first model uses word2vec or frWac as initial representations followed by Bi-LSTM or CNN classifier and a fully connected layer with a softmax activation function for prediction as shown in Figure~\ref{fig1}. The second model uses M-BERT as initial representation followed by Bi-LSTM or CNN classifier and a fully  connected layer with a softmax activation function for prediction as shown in Figure~\ref{fig2}. The different hyper-parameters including number of filters, number of epochs and batch size for each embedding and classifier achieving the best performances are shown in Table~\ref{table5}. The number of epochs and the Batch size are equal to 3 and 16, respectively. The number of filters is equal to 200 for Word2vec representation. The best performances of frWac representation and M-BERT embeddings combined with CNN are achieved when the number of filters is equal to 100. 

\begin{table}
\centering
\caption{Hyper-parameters used during the training phase.}\label{table5}
\begin{tabular}{|l|l|l|l|l|}
\hline
Embedding & Classifier & number of Filters & number of Epochs & Batch size \\
\hline
Word2vec & CNN & 200 & 3 & 16 \\
frWaC  & CNN & 100 & 3 & 16 \\
M-BERT & CNN & 100 & 3 & 16 \\
M-BERT & Bi-LSTM & - & 3 & 16 \\
\hline
\end{tabular}
\end{table}

\section{Results and Discussion}
All experiments were trained and tested on the two datasets presented in Table~\ref{table3}. 
Datasets were divided randomly for train and test without using any kind of data pre-processing technique. Models are trained according to the hyper-parameters in Table~\ref{table5}. \newline
The performance of sentiment analysis models is usually evaluated based on standard metrics adopted in the state-of-the-art. They are : Accuracy referred in our results as ACC. and the F-score. 

\begin{itemize}

\item {Accuracy} is the traditional way to measure the performance of a system.\ It represents the percentage of instances predicted correctly by the model for all class categories.

\item {F-score} provides a measure of the overall quality of a classification model as it combines the precision and recall through a weighted harmonic mean.

For our experiments two variants of F-score were used: the micro-averaging biased by class frequency (F1. micro)  and  macro-averaging taking all classes as equally important (F1. macro). This is important in the case of the TSAC-TUNIZI where classes are unbalanced.

\end{itemize}

\subsection{TUNIZI Dataset Results}
Table \ref{table6} reviews the sentiment classification
performances for TUNIZI dataset. The word2vec representation trained on the TUNIZI dataset did not achieve better performances than frWaC representation trained on a French dataset. Indeed frWac representation achieved a 70,2\% accuracy performance when word2vec representation was only 67,2\%. This could be explained by the limited size of the TUNIZI dataset (82384 words) used to train word2vec representation compared to the French dataset (1.6 billion words) used to train frWac representation. Training word2vec on the limited size of the TUNIZI dataset does not include all the Tunisian words, hence the out of vocabulary (OOV) phenomenon. The pretrained French word2vec did not solve the problem of OOV, but since it was trained on a very large corpus it handled most of French words that are frequent in Tunizi comments.

Table~\ref{table6} results suggest the outperformance
of the M-BERT embeddings over those generated by word2vec and frWac. Indeed, M-BERT embeddings combined with CNN or Bi-LSTM performed better than word2vec embeddings for all performance measures. This could be explained by the ability of Multilingual BERT to overcome the problem of switch-coding comments presented previously in Table~\ref{table4} since it includes more than 10 languages like English, French, and Spanish.  

BERT tokenizer was created with a WordPiece model. First, it will check if the whole word exists in the vocabulary, if not, then it will break down the word into sub-words and repeat the process until it becomes divided into individual characters. Because of this, a word will always have its representation in the BERT embedding matrix, and the OOV phenomenon is partially overcome. 

\begin{table}
\centering
\caption{Tunizi Classification Results.}\label{table6}
\begin{tabular}{|l|l|l|l|l|l|}
\hline
Embedding & Classifier & Dataset & ACC. & F1. micro & F1. macro \\
\hline
Word2vec & CNN & TUNIZI & 67,2\% & 67,2\% & 67,1\% \\
frWaC  & CNN & TUNIZI & 70,2\% & 70,2\% & 69\% \\

M-BERT & Bi-LSTM & TUNIZI & 76,3\% & 76,3\% & 74,3\% \\
M-BERT & CNN & TUNIZI & \textbf{78,3\%} & \textbf{78,3\%} & \textbf{78,1\%}  \\
\hline
\end{tabular}
\end{table}
\subsection{TSAC-TUNIZI Dataset Results}

To confirm TUNIZI classification results, experiments were also performed on the TSAC-TUNIZI dataset with CNN and Bi-LSTM since they showed better performances than word2vec embeddings.   
Table~\ref{table7} shows the results of the proposed models on the TSAC-TUNIZI dataset. We can notice that the BERT embedding combined with the CNN leads to the best performance with a 93,2\% of accuracy compared to 92,6\% scored by Bi-LSTM. This is also the case for the F1.micro and F1.macro performances with values 93,2\% and 93\%, respectively.

F1. micro and  F1. macro performances for CNN and Bi-LSTM are close even though the TSAC-TUNIZI dataset is unbalanced having dominant positive polarity.

\begin{table}
\centering
\caption{TSAC-TUNIZI Classification Results.}\label{table7}
\begin{tabular}{|l|l|l|l|l|l|}
\hline
Embedding & Classifier & Dataset & Acc. & F1. micro & F1. macro \\
\hline
M-BERT & Bi-LSTM & TSAC-TUNIZI & 92,6\% &  92,6\% & 92,5\% \\
M-BERT & CNN &  TSAC-TUNIZI & \textbf{93,2\%} & \textbf{93,2\%} & \textbf{93\%}  \\
\hline
\end{tabular}
\end{table}

\subsection{Discussion}
Experiments on TUNIZI and TSAC-TUNIZI datasets showed that the bidirectional encoder representations from transformers (BERT)  as a contextual language model in its multilingual version as an embedding technique outperforms word2vec representations. This could be explained by the code switching characteristic of the Tunisian dialect used on social media. Results suggest that M-BERT can overcome the out of vocabulary and code switch phenoma. This defines the M-BERT embeddings as expressive features of Tunisian dialectal content more
than word2vec embeddings and without need to translate Tunisian Romanized alphabet into Arabic alphabet. 

The CNN classifier performed better than the Bi-LSTM suggesting representation based Bi-LSTM did not benefit from the double exploration of the preceding and following contexts. This confirms the conclusion in~\cite{Jerbi} where LSTM representation outperformed Bi-LSTM representation.

Due to the non-existent work on Tunisian Romanized sentiment analysis, it wasn’t
possible to perform a comparison state of the art study. Nevertheless, the
obtained performances of our proposed approach was further
compared against the baseline systems that tackled
the TSAC dataset as shown in Table~\ref{table8}.
\begin{table}
\centering
\caption{Compared Classification Results.}\label{table8}
\begin{tabular}{|l|l|l|l|}
\hline
Model &  Dataset & ACC & F-measure (\%)  \\
\hline
~\cite{medhaffar-etal-2017-sentiment} &  TSAC & 78\%  & 78\% \\
 ~\cite{syntax} & TSAC & 86.5\% & 86.2\%   \\
~\cite{Jerbi} &  TSAC & 90\% &   -  \\
M-BERT+CNN & TSAC-TUNIZI & 93.2\% & 93\% \\
M-BERT+CNN & TSAC & \textbf{93.8\%} & \textbf{93.8\%} \\
\hline
\end{tabular}
\end{table}

Compared to the state-of-the-art applied on the tackled dataset, our results showed that CNN trained with M-BERT improved
the performance over the baselines. As we can see in Table~\ref{table8}, with the proposed approach, the performed accuracy and F-measures outperformed the baselines performances.

%

\section{Conclusion and Future work}

In this work, we have tackled the Tunisian Romanized alphabet sentiment analysis task. We have experimented two different word-level representations (word2vec and frWaC) and two deep neural networks (CNN and Bi-LSTM), without the use of any pre-processing step. Results showed that CNN trained with M-BERT achieved the best results compared to the word2vec, frWac and Bi-LSTM. This model could improve
the performance over the baselines. Experiments and promising results achieved on the TUNIZI and TSAC-TUNIZI datasets helped us to better understand the nature of the Tunisian dialect and its specificities. This will help the Tunisian NLP community in further research activities not limited to the sentiment analysis task, but also in more complex NLP tasks.

A natural future step would involve releasing TunaBERT, a Tunisian version of the Bi-directional Encoders for Transformers (BERT) that should be learned on a very large and heterogeneous Tunisia dataset. The Tunisian language model can be applied to complex NLP tasks (natural language inference, parsing, word sense disambiguation). To demonstrate the value of building a dedicated
version of BERT for Tunisian, we also plan to compare
TunaBERT to the multilingual cased version of BERT.

 \bibliographystyle{splncs04}

\begin{thebibliography}{8}

\bibitem{TAL}
Mulki, H. and  Haddad, H. and  Babaoglu, I.: 
Modern trends in Arabic sentiment analysis: A survey. Traitement Automatique des Langues \textbf{58}(3), 15--39 (2018)

\bibitem{9035299}
Guellil, I.and Azouaou, F. and Valitutti, A.: English vs Arabic Sentiment Analysis: A Survey Presenting 100 Work Studies, Resources and Tools. In: 16th International Conference on Computer Systems and Applications, pp. 1--8. IEEE Computer Society Conference Publishing Services, Abu Dhabi, UAE (2019)

  \bibitem{medhaffar-etal-2017-sentiment}
Medhaffar, S.  and Bougares, F.  and Est{\`e}ve, Y.  and
      Hadrich-Belguith, L.: Sentiment Analysis of {T}unisian Dialects: Linguistic Ressources and Experiments. In: 3rd {A}rabic Natural Language Processing Workshop, pp. 55--61. Association for Computational Linguistics, Valencia, Spain (2017)

\bibitem{10.1145/3364319}
Masmoudi, A. and Khmekhem, M. and Khrouf, M. and Hadrich-Belguith, L.: Transliteration of Arabizi into Arabic Script for Tunisian Dialect. ACM Transactions on Asian and Low-Resource Language Information Processing \textbf{19}(2), 1--21 (2019)

\bibitem{qwaider-etal-2019-modern}
Qwaider, C.  and Chatzikyriakidis, S.  and Dobnik, S.: Can Modern Standard {A}rabic Approaches be used for {A}rabic Dialects? Sentiment Analysis as a Case Study. In: 3rd Workshop on Arabic Corpus Linguistics, pp. 40--50. Association for Computational Linguistics, Cardiff, United Kingdom (2019)

\bibitem{tounsi}
Stevens , Paul B.: Ambivalence, modernisation and language attitudes: French and Arabic in Tunisia. Journal of Multilingual and Multicultural Development \textbf{4}(2-3), 101--114 (1983)

\bibitem{emerging}
Skandrani, H. and Triki, A.: Trust in supply chains,
meanings, determinants and demonstrations: A qualitative
study in an emerging market context.
Qualitative Market Research \textbf{14}(4), 391--409 (2011)

\bibitem{fourati}
Fourati, C. and Messaoudi, A. and Haddad, H.:TUNIZI: a Tunisian Arabizi sentiment analysis Dataset. \url{arXiv preprint arXiv:2004.14303}

\bibitem{younes}
Younes, J.  and  Achour, H.  and Souissi, E.: Constructing Linguistic Resources for the Tunisian Dialect Using TextualUser-Generated Contents on the Social Web. In: 15th International Conference Current Trends in Web Engineering, Rotterdam, pp. 3--14. Springer Cham, Rotterdam, The Netherlands (2015)

\bibitem{abidi}
Abidi, K.: Automatic building of multilingual resources from social networks : application to Maghrebi dialects (Doctoral dissertation). Universit{\'e} de Lorraine, France, 2019

\bibitem{survey}
Younes, J.  and Achour, H.  and Souissi, E.  and Ferchichi, A.: Survey on Corpora Availability for the Tunisian Dialect Automatic Processing. In: JCCO Joint International Conference on ICT in Education and Training, International Conference on Computing in Arabic, and International Conference on Geocomputing (JCCO: TICET-ICCA-GECO), pp. 1--7. Piscataway, NJ : IEEE, Tunis, Tunisia (2018)

\bibitem{parallelcorpus}
Bouamor, H. and Oflazer, K and Habash, N.: A Multidialectal Parallel Corpus of Arabic. In: 9th inth International Conference on Language Resources and Evaluation (LREC-2014), pp. 1240--1245. European Language Resources Association (ELRA), Reykjavik, Iceland (2014)

\bibitem{Karmani}
Karmani N.: Tunisian Arabic Customer's Reviews Processing and Analysis for an Internet Supervision System (Doctoral dissertation). Sfax University, Sfax, Tunisia, 2017

  \bibitem{sayadi}
Sayadi, K. and Liwicki, M.  and  Ingold, R.  and Bui, M.: Tunisian Dialect and Modern Standard Arabic Dataset for Sentiment Analysis : Tunisian Election Context. In: 2nd International Conference on Arabic Computational Linguistics, Konya, Turkey (2016)
 
  \bibitem{Jerbi}
Jerbi, M.  and Achour, H.  and Souissi, E.: Distributed Representations of Words and Phrases and Their Compositionality. In: 7th International
               Conference Arabic Language Processing: From Theory to Practice, pp. 122--131. Springer Cham., Nancy, France (2019)


\bibitem{haddad2018}
Mulki, H. and  Haddad, H. and  Bechikh Ali, C. and  Babaoglu, I.: Tunisian Dialect Sentiment Analysis: A Natural Language Processing-based Approach. Computaci{\'o}n y Sistemas \textbf{22}(4), 1223--1232 (2018)
\bibitem{syntax}
Mulki, H. and  Haddad, H. and  Gridach, M. and  Babaoglu, I.: Syntax-Ignorant N-gram Embeddings for Sentiment Analysis of {A}rabic Dialects. In: 4th Arabic Natural Language Processing Workshop, pp. 30--39. Association for Computational Linguistics, Florence, Italy (2019)

\bibitem{tunizi}
TUNIZI dataset, \url{https://github.com/chaymafourati/TUNIZI-Sentiment-Analysis-Tunisian-Arabizi-Dataset}. 

\bibitem{tsac}
TSAC dataset \url{https://github.com/fbougares/TSAC}
   
   \bibitem{word2vec}
Mikolov, T. and Sutskever, I. and Chen, K. and Corrado, G. and Dean, J.: Distributed Representations of Words and Phrases and Their Compositionality. In: Advances in Neural Information Processing Systems 26, pp. 3111–3119. Curran Associates Inc., Lake Tahoe, Nevada (2013)

 \bibitem{fauconnier}
French Word Embeddings, \url{http://fauconnier.github.io}. 





\bibitem{devlin-etal-2019-bert}
Devlin, J.  and
      Chang, M.  and
      Lee, K.  and
      Toutanova, K.: {BERT}: Pre-training of Deep Bidirectional Transformers for Language Understanding. In: Proceedings of the 2019 Conference of the North {A}merican Chapter of the Association for Computational Linguistics: Human Language Technologies, Volume 1 (Long and Short Papers, pp. 4171--4186. Association for Computational Linguistics, Minneapolis, Minnesota (2019)
      


  
  

    

\end{thebibliography}
%

\end{document}